\documentclass[journal]{IEEEtran}
\ifCLASSINFOpdf
\else
\fi
\usepackage{multirow}
\usepackage{threeparttable}
\usepackage{booktabs}
\usepackage{cite}
\usepackage{graphicx}  %insert graph
\usepackage{epstopdf}  %insert eps file
\usepackage{multirow}
\usepackage{array}  %assign the width
\usepackage{color}
\usepackage{colortbl}
\usepackage{amsmath}
\usepackage{amssymb}
\usepackage{mathrsfs}  %use curlicue letters
\usepackage{amsfonts}  %use hollow body letters
\usepackage[switch]{lineno}
\usepackage{algorithm}
\usepackage{algpseudocode}
\usepackage{hyperref}

\definecolor{blueColor}{rgb}{0.0,0.0,0.804}

\makeatletter
\DeclareRobustCommand\onedot{\futurelet\@let@token\@onedot}
\def\@onedot{\ifx\@let@token.\else.\null\fi\xspace}

\makeatother
\begin{document}
\title{MVP-Human Dataset for 3D Clothed Human Avatar Reconstruction from Multiple Frames}

\author{
Xiangyu Zhu$^{*}$,~\IEEEmembership{Senior Member,~IEEE},
Tingting Liao$^{*}$,
Xiaomei Zhang,
Jiangjing Lyu,
Zhiwen Chen,
Yunfeng Wang,
Kan Guo,
Qiong Cao,
Stan Z. Li,~\IEEEmembership{Fellow,~IEEE}
and Zhen Lei$^{\dag}$,~\IEEEmembership{Senior Member,~IEEE}

\thanks{
Xiangyu Zhu, Tingting Liao and Xiaomei Zhang are with Center for Biometrics and Security Research \& State Key Laboratory of Multimodal Artificial Intelligence Systems, Institute of Automation, Chinese Academy of Sciences (CASIA), Beijing
100190, China, and also with the School of Artificial Intelligence, University
of Chinese Academy of Sciences (UCAS), Beijing 100049, China (e-mail:
{xiangyu.zhu, tingting.liao, xiaomei.zhang}@nlpr.ia.ac.cn).

Jiangjing Lyu, Zhiwen Chen, Yunfeng Wang and Kan Guo are with Alibaba Group (e-mail: {jiangjing.ljj, zhiwen.czw, weishan.wyf, guokan.gk}@alibaba-inc.com).

Qiong Cao is with Speechocean(e-mail: caoqiong@speechocean.com).

Stan Z. Li is with Westlake University (e-mail: Stan.ZQ.Li@westlake.edu.cn).

Zhen Lei is Center for Biometrics and Security Research \& State Key Laboratory of Multimodal Artificial Intelligence Systems, Institute of Automation, Chinese Academy of Sciences (CASIA), Beijing 100190, China, also with
the School of Artificial Intelligence, University of Chinese Academy of
Sciences (UCAS), Beijing 100049, China, and also with the Centre for
Artificial Intelligence and Robotics, Hong Kong Institute of Science \&
Innovation, Chinese Academy of Sciences, Hong Kong (e-mail:
zlei@nlpr.ia.ac.cn).

${*}$ Equal contribution. ${\dag}$ Corresponding author
}
%and School of Artificial Intelligence, University of Chinese Academy of Sciences (Beijing 100049), China.}% <-this % stops a space
%\thanks{The corresponding author is Zhen Lei (zlei@nlpr.ia.ac.cn)}% <-this % stops a space
%\thanks{Manuscript received Aug xx, 2021.} %revised August 26, 2015.
}

\maketitle

%For pixel-level recognition, it is essential to generate features with adaptive contextual features for human parts with various sizes and shapes.

\begin{figure*}
  \includegraphics[width=0.99\linewidth]{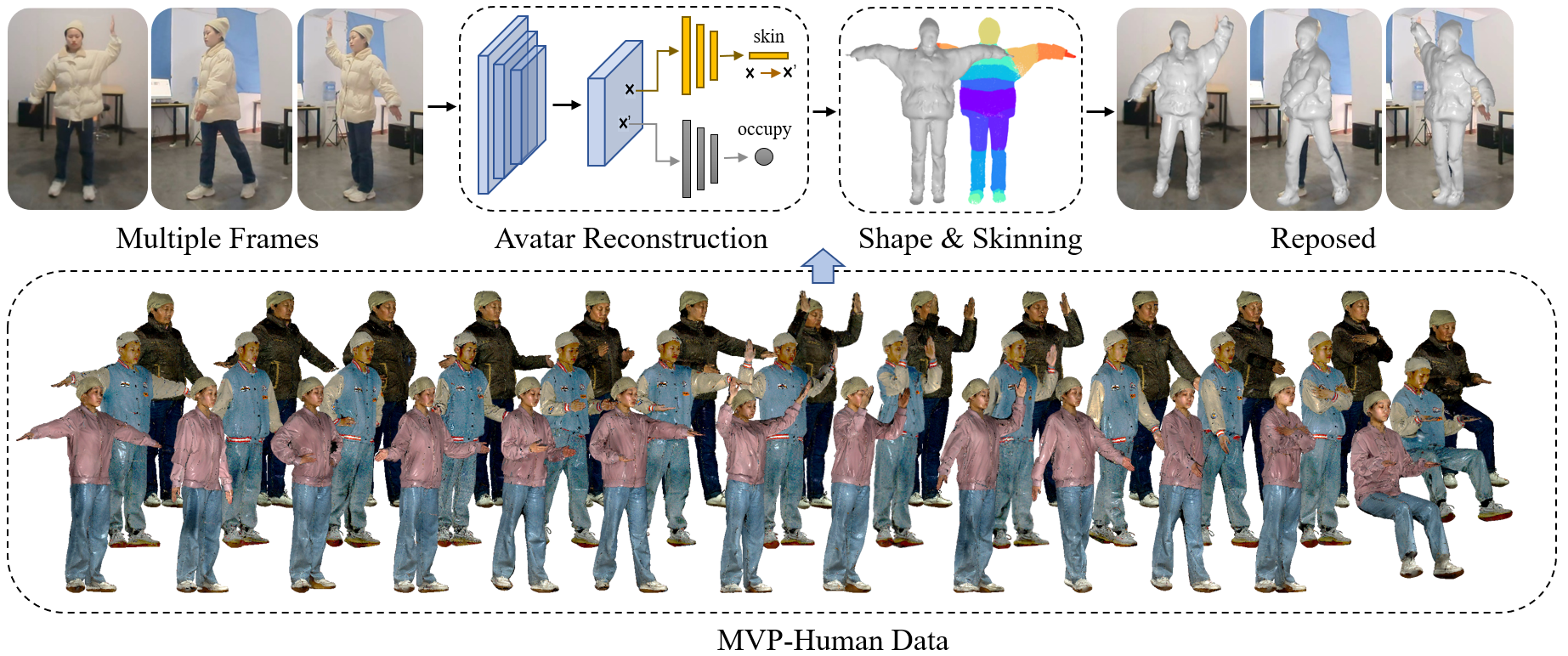}
  \caption{MVP-Human data enables the training and testing of a novel 3D avatar reconstruction model, which recovers canonical 3D shape and skinning weights of clothed humans from multiple frames in free views and poses.}
  \label{fig:teaser}
\end{figure*}

\begin{abstract}
In this paper, we consider a novel problem of reconstructing a 3D clothed human avatar from multiple frames, independent of assumptions on camera calibration, capture space, and constrained actions. We contribute a large-scale dataset, Multi-View and multi-Pose 3D human (MVP-Human in short) to help address this problem. The dataset contains $400$ subjects, each of which has $15$ scans in different poses and $8$-view images for each pose, providing $6,000$ 3D scans and $48,000$ images in total. In addition, a baseline method that takes multiple images as inputs, and generates a shape-with-skinning avatar in the canonical space, finished in one feed-forward pass is proposed. It first reconstructs the implicit skinning fields in a multi-level manner, and then the image features from multiple images are aligned and integrated to estimate a pixel-aligned implicit function that represents the clothed shape. With the newly collected dataset and the baseline method, it shows promising performance on 3D clothed avatar reconstruction. We release the MVP-Human dataset and the baseline method in \url{https://github.com/TingtingLiao/MVPHuman}, hoping to promote research and development in this field.
\end{abstract}

\begin{IEEEkeywords}
3D avatar reconstruction, 3D human, 3D body reconstruction, 3D human database.
\end{IEEEkeywords}

\section{Introduction}
\IEEEPARstart{I}{mage} based 3D human reconstruction has been widely applied in many fields such as VR/AR experience (e.g., movies, sports, games), video editing and virtual dressing~\cite{bhatnagar2019multi, santesteban2019learning, hu3DBodyNet, 9759982, zhao20183}. To reconstruct 3D humans, current methods always hold some strong assumptions about the input. 3D stereo~\cite{liu2009point, starck2007surface, wu2011shading} relies on camera calibration and a fully constrained environment. Human performance capture~\cite{habermann2019livecap, habermann2020deepcap}, which tracks human motion and deforms the 3D surface using weak multi-view supervision, needs a person-specific template before reconstruction. 3D reconstruction from single-view video fuses dynamic human appearances into a canonical 3D model, which is related to our task. However, the subject is asked to hold a rough A-pose during images collection, which
limits its applications. Besides, they either fit the SMPL model~\cite{alldieck2018detailed,alldieck2019learning} or coarsely morph the minimally-clothed body according to the silhouette~\cite{alldieck2018video}, which loses the fine-grained detail. Recently, single-image-based reconstruction~\cite{saito2019pifu,2020PIFuHD,huang2020arch,he2021arch++} learns the implicit 3D surface based on aligned image features, recovering remarkable clothes details. However, as an ill-posed problem, single-image reconstruction suffers from unsatisfactory artifacts on the back. When these methods are extended to a multi-view setting, the camera calibration is still needed~\cite{saito2019pifu,2020PIFuHD}.

To make 3D avatar reconstruction more accessible, we explore a novel problem that reconstructs a 3D avatar from unspecific frames. The 'unspecific' here means there is no requirement for camera calibration, capture space or special actions. This task is extremely challenging due to the requirement of fusing multiple snapshots in diverse views and poses into a single reconstruction. Besides, when reconstructing the geometry details, the local feature in each frame may not be reliable due to the inevitable misalignment in free poses and views, leading to overly smooth reconstruction results. Moreover, the model training relies on special training data that each person should have multi-view-and-pose images and the corresponding 3D shape in T-pose, which is inaccessible to the public.

In this work, we first create a new Multi-View multi-Pose 3D Human dataset (MVP-Human) with $400$ subjects, each having $15$ scans in different poses and $8$-view images for each pose, providing $6,000$ 3D scans and $48,000$ images in total. The linear blend skinning weights are also provided not only for supervising the model training, but also for reposing the captured T-pose meshes to the strict canonical pose, which is regarded as the target of 3D reconstruction models. Besides, benefitting from the collected multi-view images, we enable quantitative evaluation on real-world inputs, which is more reliable than the commonly used rendered meshes~\cite{saito2019pifu,2020PIFuHD,huang2020arch}.

Based on MVP-Human, we introduce a deep learning-based baseline method for 3D Avatar reconstruction from multiple frames without a camera and action specification. We propose a SKinning weights Network (SKNet) to predict the Linear Blend Skinning (LBS)~\cite{kavan2007skinning} in T-pose, by which each 3D point finds the corresponding pixel in each frame. Then the aligned image features are adaptively fused by a Surface Reconstruction Network (SRNet) to predict the final 3D shape as an implicit function~\cite{saito2019pifu}. A brief view of the framework is shown in Fig.~\ref{fig:teaser}.

The main contributions of the work mainly include two folds:
\begin{enumerate}
\item We create a large 3D clothed human dataset where high-resolution 3D scans and real images are collected in a multi-pose and multi-view setting. Sophisticated labels including 3D skeleton landmarks and linear blend skinning weights are also provided. We will make it available upon acceptance.
\item We propose a baseline method to reconstruct a clothed 3D avatar from multiple frames, independent of any camera calibration, person-specific template or specified actions. We hope the MVP-Human dataset associated with the baseline method can extend the deployment of passive 3D human reconstruction and benefit the development of this field.
\end{enumerate}

\begin{table*}
    \caption{Comparison of MVP-Human to other public clothed 3D human datasets. The column ``multi-pose scan per.'' denotes that there are multi-pose scans for each subject. The column ``multi-view images per.'' denotes that there are camera-captured multi-view images for each scan. }\label{table-datasets}
    \begin{center}
        \begin{tabular}{c|c|c|c|c|c|c}
            % \hline
             \toprule[2pt]
            Dataset & Year & \# of subjects & \# of scans   & vertices per. scan & ~multi-pose scans per.  & multi-view images per.  \\
            \midrule[1pt]
            %D-FAUST~\cite{bogo2017dynamic} & 2017 & $10$ &$40,000$& -  & - & - & - \\
            BUFF~\cite{zhang2017detailed} & 2017 & $5$ & $11,054$ & $150$k & \checkmark & - \\
            MTC~\cite{xiang2019monocular} & 2019 & $40$ & $-$ & $18.54$k & - & \checkmark \\
            Multi-Garment~\cite{bhatnagar2019multi} & 2019 & $96$ &$356$& $55$k & - & - \\
            THuman~\cite{2019DeepHuman} & 2019 & $230$ & $7,000$   & $10$k & \checkmark & -  \\
            CAPE~\cite{ma2020learning} & 2020 & $11$ & $80,000$  & $7$k & \checkmark & - \\
            HUMBI~\cite{yu2020humbi} & 2020 & $772$ & $-$ & $20$k & - & \checkmark \\
            THuman 2.0~\cite{yu2021function4d} & 2021 & $500$ & $500$ &  $289$k  & -  & -\\
            \hline
            MVP-Human (ours)& 2022 & $400$ & $6,000$  & $100$k  & \checkmark & \checkmark\\
            \hline
        \end{tabular}
    \end{center}

\end{table*}

\section{Related Work}
3D human reconstruction methods can be roughly classified by the assumptions hold on the input, including special equipment, person-specific template, and constrained actions. In this section, only the methods that take RGB inputs are discussed.

\subsection{Multi-view Stereo Reconstruction}
Multi-view acquisition approaches capture the scene from multiple synchronized cameras, and build the 3D shape from photometric cues. Stereo based methods \cite{furukawa2009accurate, waschbusch2005scalable, zitnick2004high, starck2007surface,chen2021towards} achieve remarkable performance by optimizing multi-view stereo constraints from a large number of cameras. The geometric detail can be further improved by active illumination~\cite{vlasic2009dynamic, liang2019shape, wu2012full}. However, its hardware configuration is inaccessible to general consumers due to its special equipment and complexity. Benefit from deep learning, recent works \cite{choy20163d, ji2017surfacenet, kar2017learning, gilbert2018volumetric, huang2018deep} reduce the camera number to very sparse views.
Yi et.al~\cite{yu2022multiview} propose a novel
multiview method for human pose and shape reconstruction that scales up to an arbitrary number of uncalibrated camera views (including the single view), guided by dense keypoints.
StereoPIFu \cite{yang2021stereopifu} introduces a stereo vision-based network fusing multi-view images to extract voxel-aligned features. However, this work focuses on statistical reconstruction from multi-view images in the same pose. Compared with these methods, our method can be applied to reconstruct 3D avatars from monocular frames in free views and poses without camera calibration.

\subsection{Human Performance Capture}
Most of the methods assume a pre-scanned human template and reconstruct dynamic human shapes by deforming the template to fit the images. Prior methods~\cite{de2008performance,collet2015high,starck2007surface} require a number of multi-view images as inputs, and align the template to observations using non-rigid registration. MonoPerfCap~\cite{2017MonoPerfCap} and LiveCap~\cite{habermann2019livecap} enable posed body deformation from a single-view video by optimizing the deformation to fit the silhouettes. Recently, DeepCap~\cite{habermann2020deepcap} employs deep learning to estimate the skeletal pose and non-rigid surface deformation in a weakly supervised manner, constrained by the multi-view keypoint and silhouette losses. DeepMultiCap~\cite{zheng2021deepmulticap} further extends DeepCap to the multi-person scenario.
Template-based methods need a pre-scanned template and thus focus on pose tracking and deformation, and template-free capture methods aim to estimate poses and reconstruct surface geometry at the same time. However, most approaches \cite{zheng2018hybridfusion, allain2015efficient, leroy2017multi, guo2017real, innmann2016volumedeform} are based on depth sensors which provide surface information for reconstruction, and these methods have difficulty in working on outdoor scenarios. Recently, some approaches \cite{saito2019pifu, 2020PIFuHD, he2020geo, pamir2020, xiu2022icon} utilize the technique of implicit function to generate high-fidelity meshes from single or multi-view RGB images. However, the reconstructed results can not be articulated and applied in scenarios where an avatar is needed. We introduce a template-free method which reconstructs an avatar from merely unconstrained RGB frames.
Our method shares the same input formulation, single-view RGB video, with some of these methods, but we do not need a pre-scanned template.

\subsection{3D Avatar Reconstruction}
Early works \cite{2017Detailed, 2015Dyna, 2016Estimation} employ statistical body models such as SCAPE \cite{2005SCAPE} or SMPL \cite{2015SMPL} to recover human models which are typically animatable. However, parametric models cover limited shape variations due to the restriction of PCA shape space. Recently, some methods try to recover a clothed 3D avatar from a monocular video in which a person is moving. Alldieck et.al \cite{alldieck2018video} ask the target subject to turn around in front of the camera while roughly holding the A-pose. Then the dynamic human silhouettes are transformed into a canonical frame of reference, where the visual hull is estimated to deform the SMPL model. This method is further improved by \cite{alldieck2018detailed} on fine-level details reconstruction by introducing more constraints like shape from shading. Octopus~\cite{alldieck2019learning} encodes the images of the person into pose-invariant latent codes by deep learning, and fuses the information to a canonical T-pose shape, achieving faster prediction. Such methods share the same goal of reconstructing canonical-pose shape from monocular video with our method, but we do not require the subject to perform specific actions.

Recently, neural-network-based implicit function~\cite{chen2019learning} is introduced to represent 3D humans, demonstrating significant improvement in representation power and fine-grained details than voxel~\cite{varol2018bodynet} and parametric models~\cite{2015SMPL}. PIFu~\cite{saito2019pifu} aligns individual local features at the pixel level to regress the implicit field (inside/outside probability of a 3D coordinate), enabling the 3D reconstruction from a single image. PIFuHD~\cite{2020PIFuHD} extends PIFu to a coarse-to-fine framework so that a higher resolution input can be leveraged for achieving high fidelity. MonoPort\cite{li2020monocular} proposes a faster rendering method to accelerate inference time. Yang et.al~\cite{yang2021s3} represent the pedestrian’s shape, pose, and skinning weights as neural implicit functions that are directly learned from data.
PHORHUM~\cite{alldieck2022photorealistic} presents an end-to-end methodology for photorealistic 3D human reconstruction given just a monocular RGB image.
FloRen~\cite{shao2022floren} initially recovers a coarse-level implicit geometry, which is then refined using a neural rendering framework that leverages appearance flow.
ARCH~\cite{huang2020arch,he2021arch++} reconstructs a human avatar in the canonical pose from a single image, where the image features are aligned by a semantic deformation field. These methods hold the minimum assumptions, one unconstrained image, on the input, and show state-of-the-art performance, but they suffer from depth ambiguity and unsatisfactory artifacts on the back.  It is worth noting that, PIFu can be extended to multiple views~\cite{saito2019pifu,2020PIFuHD}, but camera calibration is still needed in the scenario.

NeRF has attracted much attention in learning scene representations for novel view synthesis from 2D images. Recently, some methods introduce NeRF in 3D human reconstruction. NeRF--~\cite{wang2021nerf} aims to learn a neural radiance field without known camera parameters. AvatarGen~\cite{zhang2022avatargen} and Zheng et al.~\cite{zheng2022structured} incorporate NeRF with the human body prior to enabling animatable human reconstruction. FLAG~\cite{aliakbarian2022flag} focuses on the pose generation task for avatar animation from sparse observation by developing a flow-based generative model. Marwah~\cite{Marwah-2019-112859} presents a  two-stream methodology to infer both the texture and geometry of a person from a single image and combines the outputs of the two streams using a differentiable renderer.

\subsection{3D Human Data}
Several works have released their 3D clothed human data, summarized in Tab.~\ref{table-datasets}. Multi-Garment~\cite{bhatnagar2019multi} and THuman 2.0~\cite{yu2021function4d} release a large number of scans with various body shapes and natural poses, which have been well employed in single-view and multiview reconstructions~\cite{he2020geo,2019DeepHuman}. However, the single scan for each subject is not suitable for our task that depends on multi-pose scans. BUFF~\cite{zhang2017detailed} and CAPE~\cite{ma2020learning} capture 4D people performing a variety of pose sequences, which are compatible with
our task but the subject number is rather limited. THuman~\cite{2019DeepHuman} consists of $200$ people in rich poses, but the captured 3D meshes lose fine-grained geometry with only $10$k points.
HUMBI~\cite{yu2020humbi} is a large multiview image dataset of the human body with various expressions. However, the HUMBI dataset only published minimally-clothed SMPL models while the real scans are not released. Instead, our dataset has released both clothed 3D meshes and skinning weights for users to train avatar reconstruction models. MTC~\cite{xiang2019monocular} provides a social interaction dataset but only image sequences and 3D skeletons are available.
Compared with existing datasets, MVP-Human demonstrates particularity by its multi-pose scans of hundreds of people, high-quality scans ($10k$ vertices) and unique multiview real (not rendering) images.

\begin{figure*}
  \centering
   \includegraphics[width=0.99\linewidth]{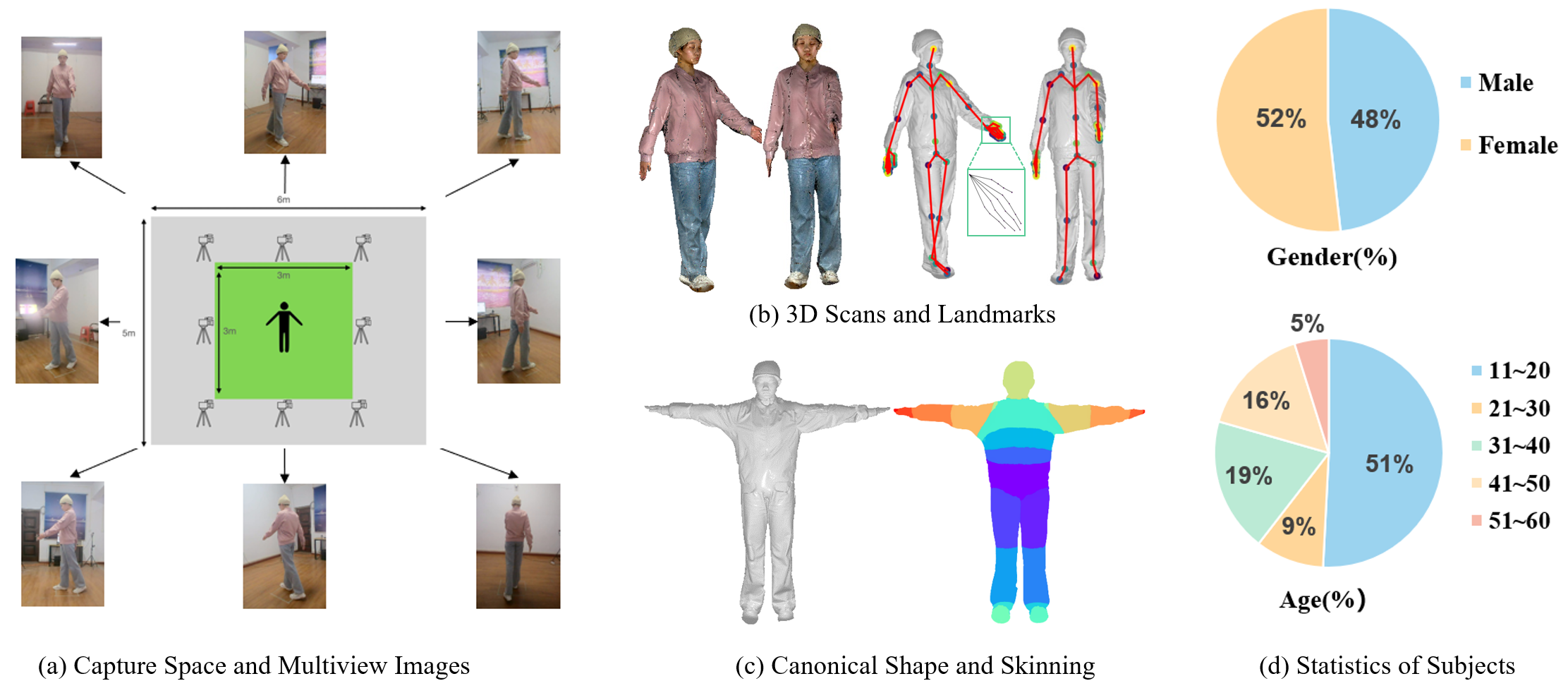}
    \caption{Multi-View multi-Pose 3D Human dataset (MVP-Human). (a) Capture space, camera placement and captured multi-view images. (b) Textured 3D scan and labeled 3D landmarks. (c) Optimized canonical mesh and skinning weights which serve as the target of avatar reconstruction. (d) The statistics of subjects in MVP-Human.}
     \label{fig:setup_and_sample}
\end{figure*}

\section{Multi-pose Multi-view 3D Human Dataset}
\label{sec:data}
Reconstructing a 3D avatar from unspecific frames usually relies on the 3D textured scans of one subject in rich poses and its T-pose canonical mesh with skinning weights, which cannot be provided by existing datasets. Among existing datasets in Table.~\ref{table-datasets}, BUFF \cite{Zhang_2017_CVPR}, CAPE~\cite{ma2020learning}, and THuman~\cite{2019DeepHuman} are related datasets that provide multi-pose meshes for each subject, but the models in CAPE and THuman have about $7$k and $10$k 3D points, respectively, which are insufficient to model cloth details. Besides, the BUFF has limited subjects which are insufficient to train a generalized neural network. Besides, we wish to improve quantitative evaluation from rendered synthetic images to realistic inputs, requiring real-world multi-view and multi-pose images, and the corresponding ground-truth 3D shape. To this end, we create a Multi-View multi-Pose 3D Human dataset (MVP-Human), a large 3D human dataset containing rich variations in poses and identities, together with the multi-view images for each subject.

\subsection{Data Capture}
\begin{figure*}
  \centering
   \includegraphics[width=0.99\linewidth]{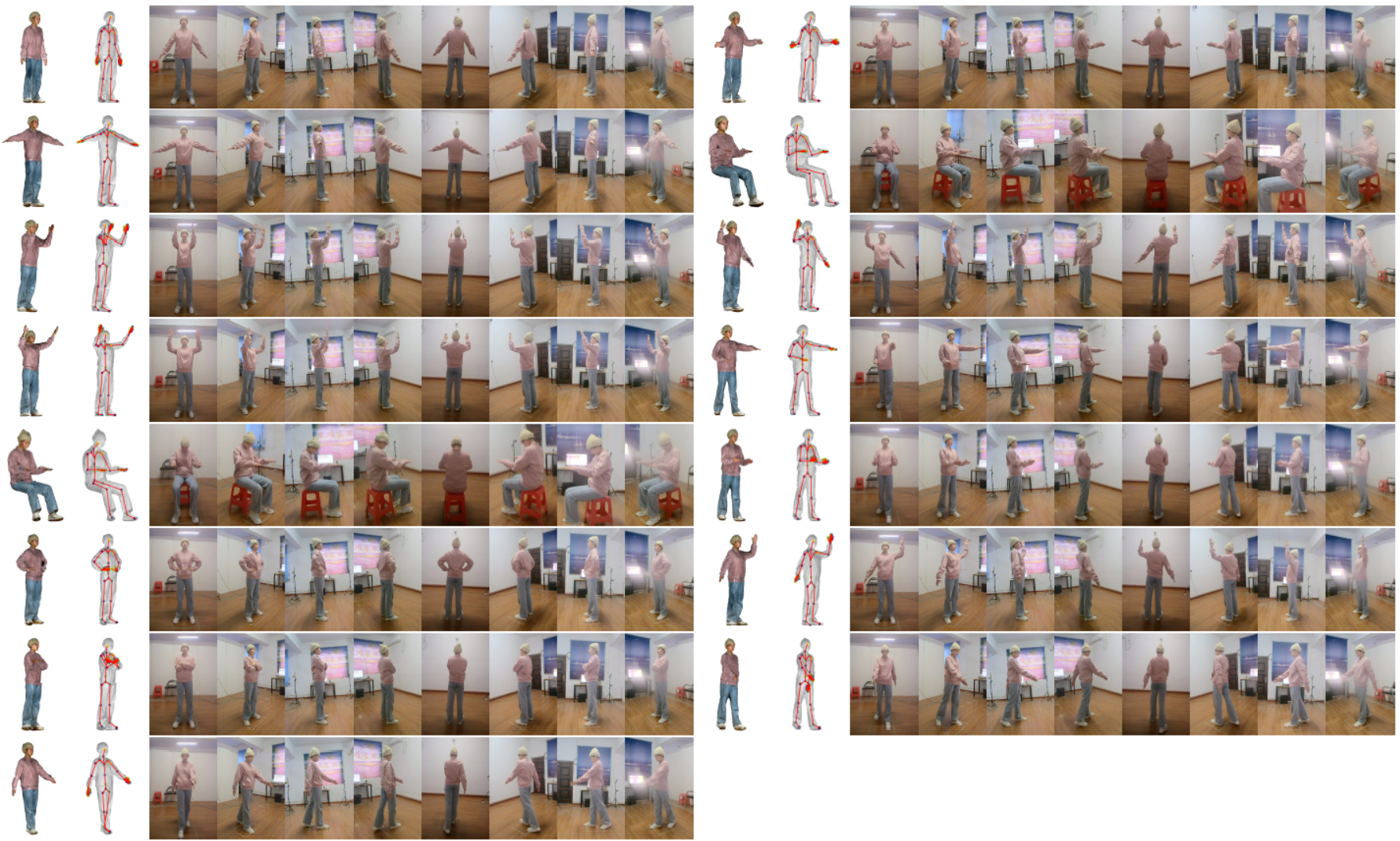}
    \caption{The collected data for a subject in MVP-Human. Each subject is captured in $15$ poses, and each pose has a 3D textured scan (first column),  annotated 3D landmarks (second column), and multi-view images (other columns). The images are cropped for better visualization.}
     \label{fig:fig-subject_imgs}
\end{figure*}

\noindent \textbf{Capture Setting}: The capturing of the dataset took place in a custom-built multi-camera system. As shown in Fig.~\ref{fig:setup_and_sample}(a), the venue was $5$m $\times$ $5$m, and within it we obtained a capture space of $3$m $\times$ $3$m, where subjects were fully visible in all video cameras. Totally $8$ capture components were placed around the area at equal $45^{\circ}$ intervals to give full-body capture for a wide range of motions. Each component consisted of one Shenzhen D-VITEC  RGB camera of $1080 \times 720$ resolution, $90^{\circ}$ FOV, and $30$ FPS for image collection, and one Wiiboox Reeyee Pro 2X 3D body scanner for 3D capture. The outputs of the system were multi-view $1080 \times 720$ images and 3D meshes with approximately $100,000$ vertices, $2,000,000$ faces, and $8400\times8400$ texture maps. Then, 3D landmarks with $62$ points are manually labelled for each 3D mesh (Fig.~\ref{fig:setup_and_sample}(b)). The 3D landmarks are the combination and extension of MPII~\cite{andriluka20142d} human body landmarks and FreiHAND~\cite{zimmermann2019freihand} hand landmarks. Specifically, the 62-landmarks contain $16$ MPII landmarks, $40$ FreiHand landmarks (without the wrist), and additional $6$ landmarks on the upper chest, chest, left upper arm, right upper arm, left toe, and right toe defined by ours. The 3D landmarks are annotated by well-trained Maya engineers, and all the results are checked by a quality inspector to ensure consistency.
Besides, optimized canonical mesh and skinning weights serve as the target of avatar reconstruction (Fig.~\ref{fig:setup_and_sample}(c)). More samples are demonstrated in Fig.~\ref{fig:fig-subject_imgs}, where each subject is captured in $15$ poses, and each pose has a 3D textured scan (first column),  annotated 3D landmarks (second column), and multi-view images (other columns).

\begin{figure*}
  \centering
   \includegraphics[width=0.95\linewidth]{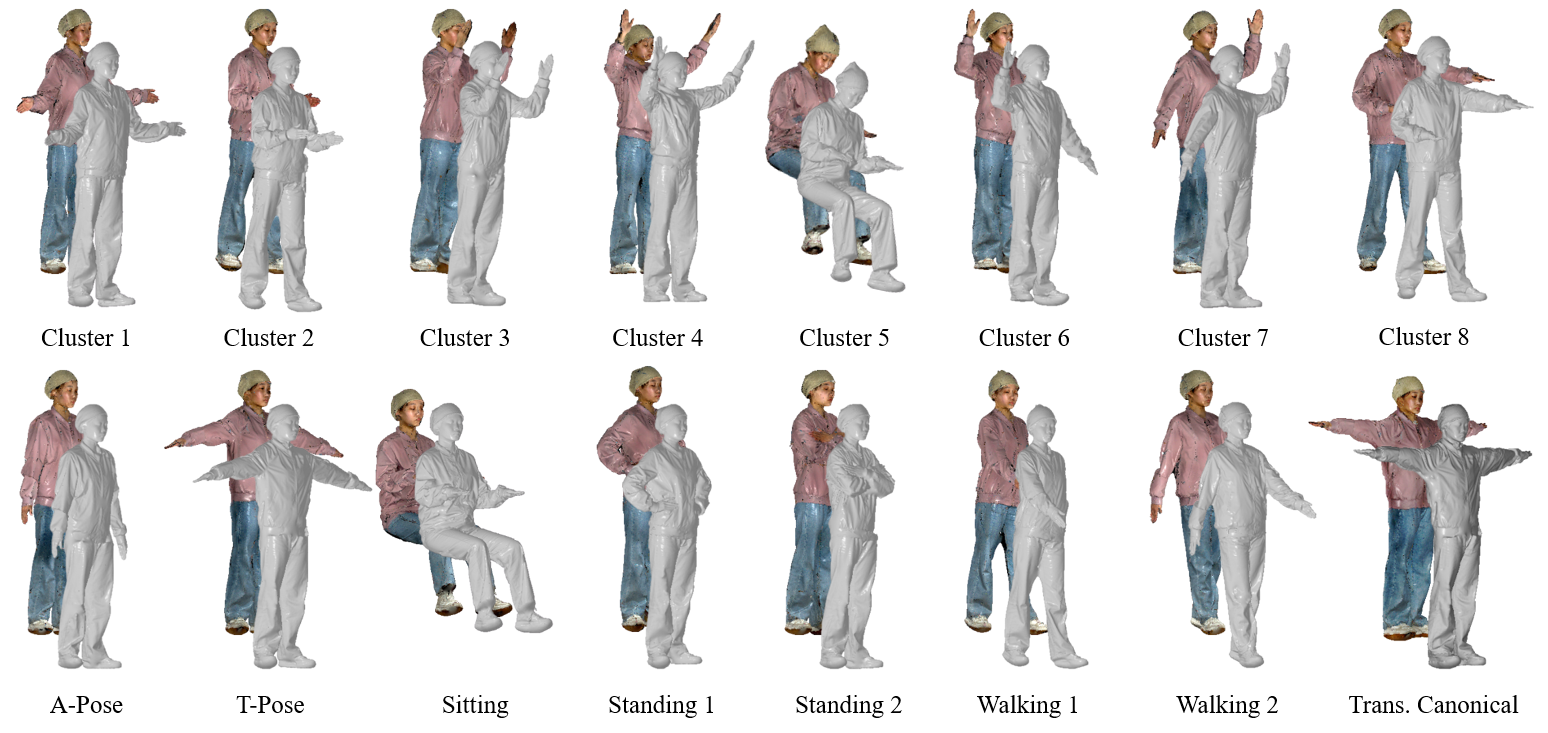}
   \caption{Specified $15$ poses and canonical pose (strict T-pose) in MVP-Human. The specified poses consist of the clustered AMASS~\cite{mahmood2019amass} poses and several common poses, such as T-pose, sitting, and walking.}
   \label{fig:actions}
\end{figure*}

\noindent \textbf{Subjects and Poses}: For the creation of the dataset, $400$ distinctive actors with balanced gender and age were recruited. The statistics are summarized in Fig.~\ref{fig:setup_and_sample}(d). Most of the subjects wore their own clothes, and some of them were asked to wear prepared clothes like common jackets, pants, and tracksuits to maintain realism. This choice provides rich variability in body shape and mobility. Besides, to reflect the common actions in daily life, we obtained $8$ typical poses by clustering the SMPL pose parameters of AMASS~\cite{mahmood2019amass} and added $7$ common poses, including T-pose, A-pose, sitting, standing with arms akimbo, standing with crossed arms, and two walking poses, which did not emerge in the clustering results. Finally, we get $15$ poses as in Fig.~\ref{fig:actions}.

Generally, the MVP-Human provides $15$ poses and $8$-view images for $400$ subjects, proving $6,000$ 3D scans and $48,000$ images in total. To the best of our knowledge, MVP-Human is the largest 3D human dataset in terms of the number of subjects along with high-quality 3D body meshes and multi-view RGB images.

\begin{figure*}[t]
\begin{center}
  \includegraphics[width=0.99\linewidth]{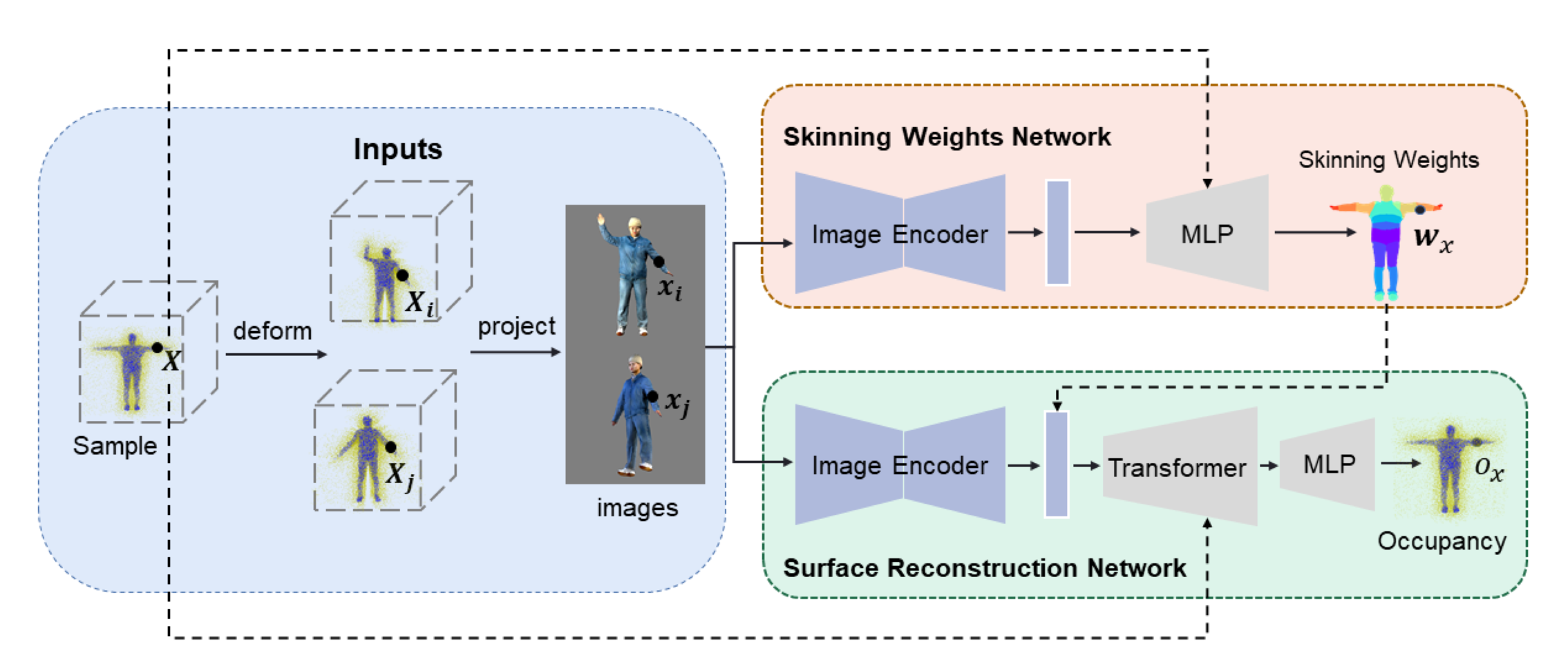}
\end{center}
  \caption{Overview of our baseline framework. The SKinning weights Network (SKNet) estimates the skinning weights from an image, which are leveraged by the Surface Reconstruction Network (SRNet) to predict the occupancy field of a clothed human. }
  \label{fig-framework}
\end{figure*}

\subsection{Skinning Weights Construction}
\label{sec:data-skin}
To create the linear blend skinning weights as the supervision of SKNet, we optimize the canonical 3D model with the skinning weights for each subject following the recent SCANimate \cite{saito2021scanimate}, shown in Fig.~\ref{fig:setup_and_sample}(c). Under the support of MVP-Human, where every subject has multiple scans in different poses with labeled 3D landmarks, we do not employ the original circle consistency loss and modify SCANimate in two aspects: 1) we directly optimize the chamfer errors brought by reposing, calculated by the distances between the reposed shape and the target-pose scan. 2) The reposing is guided by hand-labeled 3D landmarks rather than detected landmarks. The advanced SCANimate provides more reliable skinning weights, which are not only utilized as the target of SKNet but also used to repose the scanned T-pose shape to the strict T-pose as the target of SRNet. We also cut the edge in self-intersection regions according to \cite{saito2021scanimate} and fill the hole using smooth signed distance surface reconstruction \cite{calakli2011ssd}.

%\subsection{Authorization}
%The MVP-human will be released to the public for academic use only, under the licence following the Restrictive Licence Template\footnote{https://library.unimelb.edu.au/Digital-Scholarship/restrictive-licence-template}, which is specially designed for privacy protection. All the actors have signed the authorization letters (an example is provided in supplemental materials) that approve us to release their 3D mesh and image data.

\section{Avatar Reconstruction Baseline with MVP-Human Dataset}\label{sec:method}
We first present our method and describe its dependence on our data. Our goal is to estimate the T-pose shape in the canonical space from multiple frames without constraint on camera, space, and actions. We employ the implicit function~\cite{saito2019pifu} as the 3D representation to formulate the task as:
\begin{equation}\label{equ-overview}
  f(F(\mathbf{x}_{1}, \mathbf{I}_{1}), ..., F(\mathbf{x}_{V}, \mathbf{I}_{V} ),\mathbf{X})=o:o \in \mathbb{R},
\end{equation}
where for a 3D point $\mathbf{X}$ in the canonical space, $x_{v}$ is its 2D perspective projection on the $v$th image $\mathbf{I}_{v}$, $F(\mathbf{x}_{v}, \mathbf{I}_{v})$ is the corresponding local image feature and $o$ is the  inside/outside probability of the 3D point.
The camera poses for projection are estimated together with the human body poses during SMPL fitting.
In this section, we propose a baseline method equipped with two sub-networks to achieve this goal: a skinning weights network that learns the blend skinning weights for pixel alignment, and a surface reconstruction network that fuses image and geometry features to a canonical 3D shape.

\subsection{Pixel Alignment via Skinning Weights}
PIFu~\cite{saito2019pifu} has demonstrated that the pixel-aligned image feature is the key to reconstructing detailed 3D surfaces.
In posed 3D human reconstruction~\cite{saito2019pifu,2020PIFuHD}, the image feature of a 3D point $(x,y,z)$ can be naturally retrieved on the $(x,y)$ of the image plane. However, in avatar reconstruction, the retrieval of 2D projected coordinates $\mathbf{x}_{1}, \mathbf{x}_{2}, ..., \mathbf{x}_{V}$ for the 3D point $\mathbf{X}$ is not straightforward due to the unknown views and poses of the input frames.

To rebuild the canonical-to-image correspondence, we employ a skeleton-driven deformation method, as shown in Fig.~\ref{fig-framework}. Specifically, each 3D vertex on the body surface can be transformed to any pose using a weighted influence of its neighboring bones, defined as $\mathbf{w} = \{w_{k}\}_{k=1}^{K}$ on $K$ joints $\mathbf{J}=\{j_{k}\}_{k=1}^{K}$, called Linear Blend Skinning (LBS)~\cite{kavan2007skinning}. For each 3D surface point $\mathbf{X}$ in the canonical space, given the target pose parameters $\mathbf{R}$ represented by relative rotations and the projection matrix $\mathbf{P}$ calculated by the camera external parameters, its projected location is:
\begin{equation}\label{equa:LBS}
  \mathbf{x} = \mathbf{P} * \left(\sum_{k=1}^{K} w_{k}\mathcal{G}_{k}(\mathbf{R}, j_{k})\right) * \mathbf{X},
\end{equation}
where $w_{k}$ is the skinning weight and $\mathcal{G}_{k}(\mathbf{R}, j_{k})$ is the affine transformation that transforms the $k$th joint from the canonical pose to the target pose~\cite{huang2020arch}. Thanks to the development of human model fitting, we can estimate the pose $\mathbf{R}$, the corresponding $\mathcal{G}_{k}(\mathbf{R}, j_{k})$, and the projection $\mathbf{P}$ by fitting the SMPL to each frame~\cite{kolotouros2019spin} independently. Therefore, the main challenge is the unknown skinning weights $\mathbf{w}$, especially the 3D shape is inaccessible by now.

\begin{figure}[t]
\begin{center}
  \includegraphics[width=1.0\linewidth]{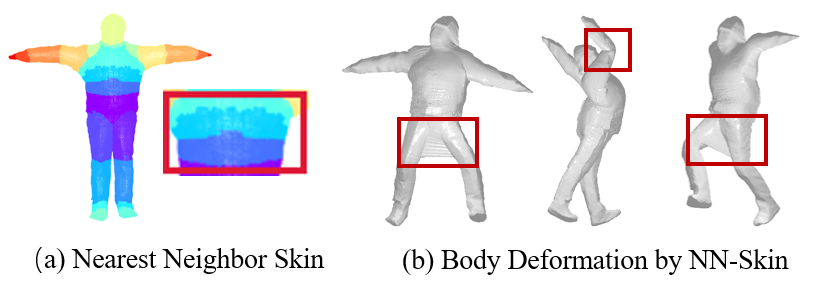}
\end{center}
  \caption{Nearest neighbor skinning weights (NN-Skin). (a) The skinning weights on the boundary of body parts are not continuous. (b) Surfaces may be ripped or adhered after deformation.}
  \label{fig:nn-skin}
\end{figure}

\noindent \textbf{Skinning Weights Network (SKNet)}: To estimate the Linear Blend Skinning (LBS) without 3D shape, we extend the definition of LBS beyond the body surface, and each $\mathbf{X}$ in the space is assigned skinning weights. Intuitively, since the regions close to the human body are highly correlated with the nearest body parts, we can place a mean-shape template in the canonical space and estimate the skinning weights according to the closest point on the template, whose skinning weights are provided by the SMPL~\cite{2015SMPL}.  This Nearest Neighbor Skinning weights (NN-Skin)~\cite{huang2020arch}, as shown in Fig.~\ref{fig:nn-skin}(a), are not reliable due to two drawbacks. First, the skinning weights on the boundary of body parts are not continuous due to the nearest-neighbor criterion. After deformation, surfaces may be ripped or adhered due to the error on the NN matching, as shown in Fig.~\ref{fig:nn-skin}(b). Second, this error can be further amplified when the mean template does not resemble the ground-truth shape.

To address this problem, we propose the SKinning weights Network (SKNet) to learn an implicit skinning field~\cite{saito2021scanimate} in the canonical space from multiple images. Given a 3D point $\mathbf{X}$, we build the SKNet to regress the skinning weights by:
\begin{gather}\label{eqn-sknet}
\mathbf{w}^{skin} = Net^{skin}(F(\mathbf{x}_{1}, \mathbf{I}_{1}), ..., F(\mathbf{x}_{V}, \mathbf{I}_{V}),\mathbf{X}),  \\
\mathbf{x}_{v} = \mathbf{P}_{v} * \left( \sum_{k=1}^{K} w_{k}^{nn}\mathcal{G}_{k}(\mathbf{R}_{v}, j_{k}) \right) * \mathbf{X}, ~~ v = \{1,2,...,V\} \notag,
\end{gather}
where for each 3D point $\mathbf{X}$ in the canonical space, the Network $Net^{skin}$ takes the corresponding image feature $F(\mathbf{x}_{v}, \mathbf{I}_{v})$ in each frame and regresses the skinning weights through an MLP architecture. The 2D projection on each frame $\mathbf{x}_{v}$ is achieved as in Eqn.~\ref{equa:LBS}, where the NN-Skin $w_{k}^{nn}$ is employed.
Particularly, the SKNet outputs $K+1$ ($K$ is the joint number) channels, where the first $K$ channels are the skinning weights and the last channel holds the inside/outside probability as auxiliary supervision. The label is $(\mathbf{w}^{*}, 0)$ for an inside point and $(0, 0, ..., 1)$ for an outside point. The target skinning weights are provided by the new MVP-Human dataset. Compared to NN-Skin, SKNet enables more accurate 2D projection and avoids cracks during avatar animation, which is validated in the experiments.

%\begin{comment}
%\begin{gather}\label{eqn-sknet}
%\arg \min\limits_{\theta} \| \mathcal{N}^{skin}(F(\mathbf{x}_{1}, \mathbf{I}_{1}), ..., F(\mathbf{x}_{V}, %\mathbf{I}_{V}),\mathbf{X};\theta) - \mathbf{w}^{*} \|, \notag \\
%\mathbf{x}_{v} = \mathbf{P}_{v} * \left( \sum_{k=1}^{K} w_{k}^{NN}\mathcal{G}_{k}(\mathbf{R}_{v}, J_{k}) \right) * \mathbf{X}, ~~ %v = \{1,2,...,V\}
%\end{gather}
%where for each 3D point $X$ in the canonical-pose space, the Network $Net$ with parameters $\theta$ takes the corresponding image feature $F(\mathbf{x}_{v}, \mathbf{I}_{v})$ in each frame and regresses the skinning weights $\mathbf{w}^{*}$ through an MLP architecture. The 2D projection on each frame $\mathbf{x}_{v}$ is achieved as in Eqn.~\ref{equa:LBS} where the NN-Skin $w_{k}^{NN}$ is employed. Particularly, the SKNet outputs $K+1$ ($K$ is the joint number) channels, where the first $K$ channels are the skinning weights and the last channel holds the existence. The label for an inside point is $(\mathbf{w}^{*}, 0)$ and $(0, 0, ..., 1)$ for an outside point. Compared to NN-Skin, SKNet enables more accurate 2D projection and avoids crack during transformation.
%\end{comment}

\subsection{Surface Reconstruction via Adaptive Fusion}

\noindent \textbf{Surface Reconstruction Network:}
For surface reconstruction, we follow Eqn.~\ref{equ-overview} and use the occupancy function $O$ to implicitly represent the 3D clothed human~\cite{huang2020arch}:
\begin{equation}\label{eqn-occup-fun}
    O =\{(o,\mathbf{X}): X\in\mathbb{R}^{3}, 0 \le o \le 1\},
\end{equation}
where $(o,\mathbf{X})$ denotes the occupancy value for one point $\mathbf{X}$ in the canonical space. In this paper, the $(o,\mathbf{X})$ is represented by a Surface Reconstruction Network (SRNet):
\begin{equation}\label{eqn-srnet}
    o = Net^{rec}(\mathbf{F}^{I}, \mathbf{F}^{S},\mathbf{X}),
\end{equation}
where the network $Net^{rec}$ takes the point $\mathbf{X}$, its multi-frame image feature $\mathbf{F}^{I}$, and a spatial feature $\mathbf{F}^{S}$ as inputs and estimates occupancy as the 3D shape.

\noindent \textbf{Image Feature:}
After the estimation of SKNet, the 2D projections $\mathbf{x}_{v}$ for each frame $\mathbf{I}_{v}$ can be achieved by Eqn.~\ref{equa:LBS}, and the image feature can be extracted by bilinear sampling on the feature map of the image encoder. However, under unconstrained scenarios, the image features may not share the same importance since 1) the image feature may not describe the 3D point due to self-occlusion, and 2) the fitted SMPL may provide an unreliable pose and view, leading to the sampling on the background. So directly averaging inferred features~\cite{2020PIFuHD} is improper. In this paper, we do not estimate blending weights by hand-crafted criteria but leverage the attention mechanism~\cite{vaswani2017attention} for adaptive feature fusion.

Transformer~\cite{vaswani2017attention} is originally proposed to capture the correlations across input sequences, and its self-attention mechanism is naturally extendable to our task.  We employ a multi-head attention network as~\cite{dosovitskiy2020image} to encode the relevance between features:
\begin{equation}\label{eqn-funsion}
\begin{aligned}
    \mathbf{F}^{I} = {\rm Attention}(\mathbf{t}_{1}, ..., \mathbf{t}_{V}&), \\
    \mathbf{t}_{v} = {\rm Concat}(F(\mathbf{x}_{v}, \mathbf{I}_{v}), \mathop{\mathbf{n}_{v}}\limits^{\rightarrow}&),
    \end{aligned}
\end{equation}
where $\mathbf{F}^{I}$ is the fused image feature, $\mathbf{t}_{v}$ is the input token of the transformer, and each token is the concatenation of the retrieved image feature $F(\mathbf{x}_{v}, \mathbf{I}_{v})$ and its visibility, represented by the surface normal $\mathop{\mathbf{n}_{v}}\limits^{\rightarrow}$ on the fitted SMPL. The normal can encode visibility because when the z of normal $\mathop{\mathbf{n}}\limits^{\rightarrow}$ is negative, it means the vertex faces inward and it is invisible. Otherwise, if the z is positive, the vertex is visible. We can see that the image feature $F(\mathbf{x}_{v}, \mathbf{I}_{v})$ is extracted from the top-level feature map of the image encoder. Even if the skinning weights are sub-optimal, each feature has a receptive field that implicitly aligns the error.

\noindent \textbf{Spatial Feature:}
Recent methods~\cite{huang2020arch,he2021arch++} have proved the effectiveness of the spatial feature that represents the geometric primitive of those points. In this work, we directly employ the estimated skinning weights as the spatial feature $ \mathbf{F}^{S} = \mathbf{w}^{skin}(\mathbf{X})$ since it encodes the relationship of that point to each of the body landmarks. This mechanism avoids any external computation cost like hand-crafted features~\cite{huang2020arch} and PointNet~\cite{he2021arch++}.

The training and testing of avatar reconstruction models depend on a sophisticated dataset equipped with multi-pose scans, multi-view images, and the strict T-pose shape for each subject, which is described as follows.

%directly employ a more reliable supervised loss to minimize the Chamfer error of the sampled point cloud pairs. Meanwhile, the SMPL model is fitted by the labeled 3D landmarks to provide the shape and pose parameters required by SCANimate.

\begin{figure*}[t]
\begin{center}
  \includegraphics[width=0.92\linewidth]{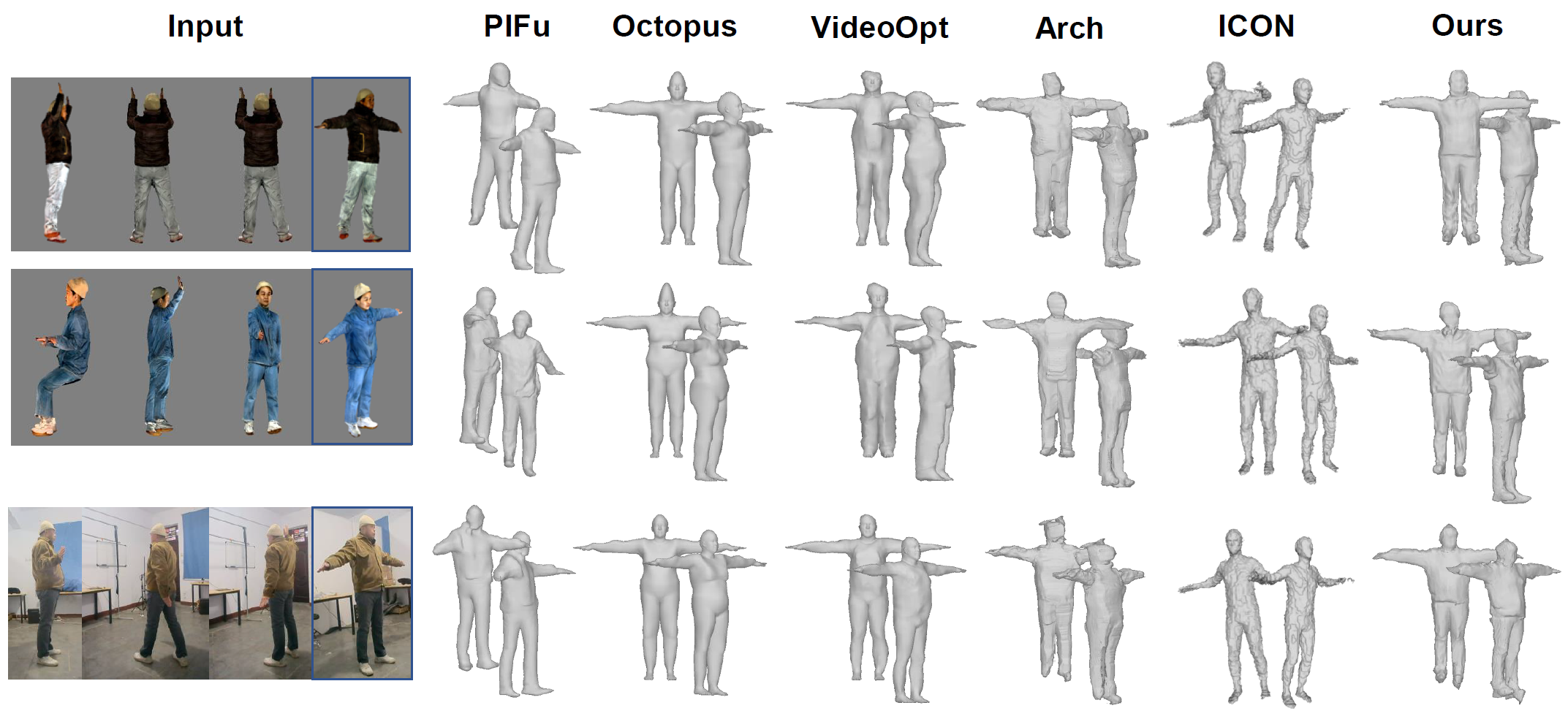}
\end{center}
  \caption{Qualitative comparison against the state-of-the-art methods. The inputs are rendering and real images of MVP-Human testing set. For each sample, the image in the bottom right (blue rectangle) is the input for PIFu and ICON.}
  \label{comparison}
\end{figure*}

\section{Experiments}
\label{experiment}

\subsection{Implementation Details}
The two sub-network follow the same MLP architecture~\cite{saito2019pifu} with the intermediate neuron dimensions of $(1024, 512, 256, 128)$ and use the Stacked Hourglass~\cite{newell2016stacked} as the image encoder. The SKNet has a $280$-dimensional input including $256$-dimensional averaged image feature and $24$-dimensional position encoding from the 3D coordinate~\cite{mildenhall2020nerf}. The output is the $24$-dimensional skinning weights plus an inside-outside probability. The input of SRNet has an additional $24$-dimension spatial features, and the output is occupancy.
The transformer for image fusion follows the architecture of~\cite{vaswani2017attention} with $4$ heads and $2$ layers, which takes $4$ image features as inputs and outputs a $256$-dimensional fused feature. During training, we first train the SKNet, whose parameters are utilized to initialize SRNet, and then jointly train the two sub-networks. The L2 loss is employed for both SKNet and SRNet. The Adam optimizer is adopted with the learning rate of $1\times10^{-3}$, decayed by the factor of $0.1$ at the $500$th epoch, and the total number of epochs is $800$. Each mini-batch is constructed by $16$ images of $4$ subjects, with $4$ images randomly selected from one subject, and $20,000$ points are sampled around the canonical mesh with a standard deviation of $0.1$ cm.

\subsection{Dataset}
We split the MVP-Human into a training set of $350$ subjects and a testing set of $50$ subjects. For the training set, we create the rendering images of 3D meshes following~\cite{saito2019pifu}, where $360$ images are produced by rotating the camera around the vertical axis with intervals of $1^{\circ}$, generating $2,100,000$ images. The ground-truth canonical shape and skinning weights are used for supervision. For the testing set, $4$ images are randomly selected from the image collection of a subject as one testing sample, which is repeated $100$ times to generate $5,000$ testing samples. It is worth noting that, this testing set first enables the quantitative evaluation of real-world images. Besides MVP-Human, we also employ People-Snapshot~\cite{alldieck2018video} in qualitative evaluation, where the subjects are asked to rotate while holding an A-pose, whose scenario can be considered as an easier case of ours.

\subsection{Comparisons}

We evaluate the proposed baseline method for clothed 3D avatar reconstruction from multiple unspecific frames on MVP-Human dataset. To perform the comparison, we introduce a method solving a highly related task, and a modification of a state-of-the-art method.

\textit{VideoOpt (Video based Optimization)}~\cite{alldieck2018video} takes a monocular RGB video where a subject rotates in an A-pose and generates the SMPL model with per-vertex offset by aligning rendered silhouettes to observations. This method can be  generalized to the inputs without an A-pose assumption but suffers from performance deterioration due to complicated silhouettes in the unconstrained environment.

\textit{Octopus}~\cite{alldieck2019learning} fits the SMPL+D model from semantic segmentation and 2D keypoints. Similar to VideoOpt, Octopus also employs the shape offsets to represent personal details, such as hair and wrinkles, but is much faster by employing a deep learning model.

\textit{ARCH} is denoted as a multi-frame extension of the state-of-the-art method ARCH~\cite{huang2020arch}. The original ARCH learns to reconstruct a clothed avatar from a single image, which is extended by us to take multiple images. The main difference between ARCH and our baseline method is that we try to estimate skinning weights rather than approximating that by the nearest neighbor, and we fuse the image features by a sophisticated model to adapt to the complicated unconstrained environment.

\textit{PIFu~\cite{saito2019pifu}} is a single-image reconstruction method that is designed to reconstruct the human body in its original posed space. However, it faces challenges in handling the MVP-Human dataset due to the lack of prior knowledge about the human body. The error is large because PIFu cannot reconstruct strict T-pose mesh as other avatar reconstruction methods.

\textit{ICON~\cite{xiu2022icon}} performs better than PIFu due to its incorporation of prior knowledge about the human body. However, the absence of a global image encoder in ICON leads to a reduction in surface smoothness, resulting in a slightly bumpy surface for the reconstructed 3D avatar.
% \textit{ARCH++}~\cite{he2021arch++} further extends ARCH by incorporating PointNet++~\cite{qi2017pointnet, qi2017pointnet++} based semantic aware
% geometry encoder to learn the underlying 3D human body prior, so that the self-contact problem of ARCH can be reduced. Similar with ARCH, ARCH++ also uses NN skinning weight as well. Since ARCH++ is not open source, we reproduce its code.

\begin{table}
    \centering
    \caption{Quantitative comparisons of normal, P2S (cm), and Chamfer error (cm) on the MVP-Human, evaluated on the canonical shape. Lower values are better.}
      \resizebox{0.40\textwidth}{!} {
      \begin{tabular}{cccc}
      \toprule[2pt]
      Method& Normal & P2S & Chamfer   \\
      \midrule[1pt]
      PIFU~\cite{saito2019pifu} & 0.8013  & 6.1931   & 4.9638       \\
      ICON~\cite{xiu2022icon} & 0.1957  &3.9583    &3.9583        \\
      Octopus~\cite{alldieck2019learning}  &  0.0266 & 2.2017  & 2.4854       \\
      VideoOpt~\cite{alldieck2018video}&  0.0233  & 1.6590  & 1.9960        \\
      ARCH~\cite{huang2020arch} & 0.0217  &  1.4208   & 1.8064       \\

      \textbf{Ours} &  \textbf{ 0.0214 } & \textbf{ 1.3418  } & \textbf{ 1.7192 }   \\
      \bottomrule[1.5pt]
      \end{tabular}
      }
    \label{tab-compare}
\end{table}

\begin{figure}[t]
\begin{center}
  \includegraphics[width=0.95\linewidth]{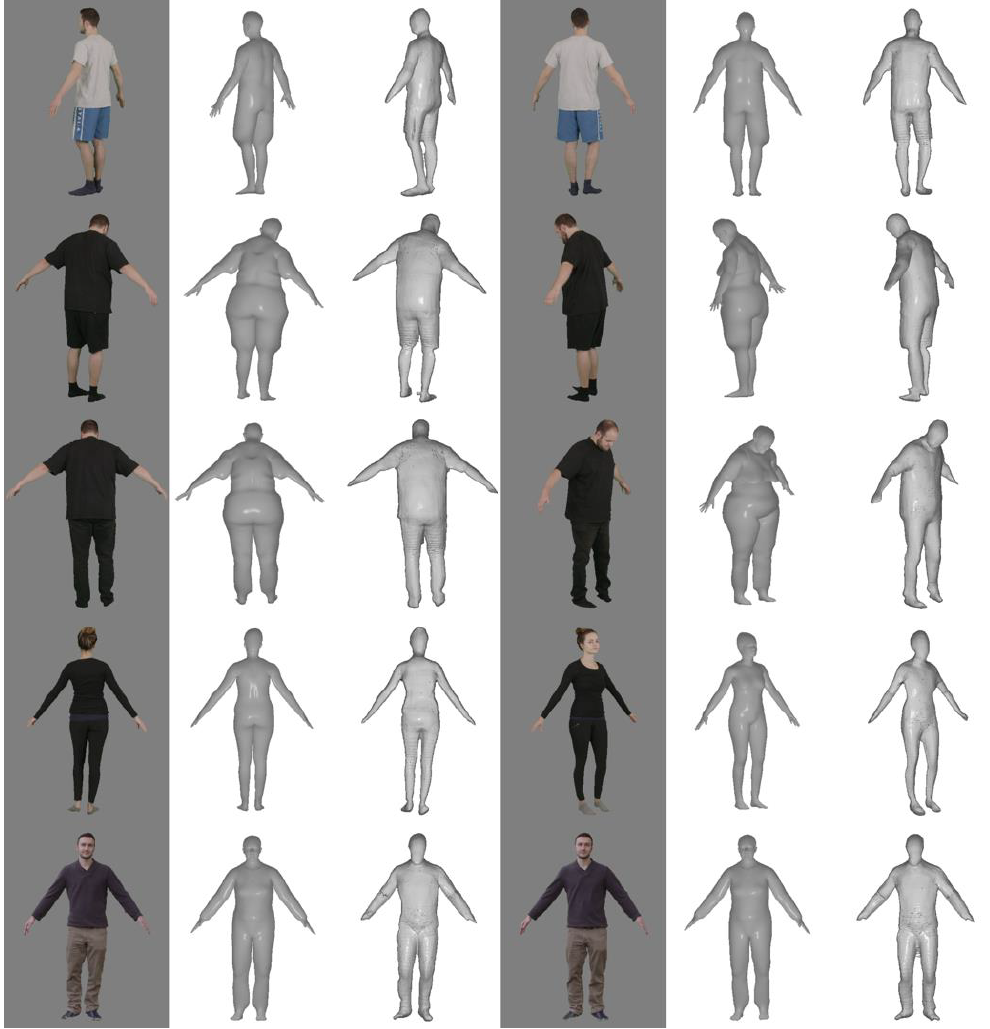}
\end{center}
  \caption{Qualitative comparison on the People-Snapshot dataset. Each sample shows one of the inputs (left), the result of VideoOpt~\cite{alldieck2018video} (middle), and our result (right). Besides the accuracy, our method is much faster than VideoOpt since only $4$ frames are utilized.}
  \label{snapshot}
\end{figure}

For quantitative comparison, we evaluate the accuracy of the reconstructed canonical 3D shape on the testing set of MVP-Human (real images), with three metrics similar to \cite{saito2019pifu}, including the normal error, point-to-surface (P2S) distance, and Chamfer error. For fairness, the ARCH is fine-tuned on the MVP-Human training set. As shown in Table~\ref{tab-compare}, we achieve the best results.

For qualitative comparison, Fig.~\ref{comparison} shows the reconstructed shapes from rendering and real images of MVP-Human, and Fig.~\ref{snapshot} shows the results in People-Snapshot, respectively. It can be seen that Octopus~\cite{alldieck2019learning} and VideoOpt~\cite{alldieck2018video}, which attempt to learn SMPL offset, have difficulty in capturing fine-grained geometry due to the limited representation power of the shape space. ARCH retains some distinctive shapes but suffers artifacts like discontinuous surfaces and distorted body parts, which come from misaligned image features.

\begin{figure}[t]
\begin{center}
  \includegraphics[width=0.95\linewidth]{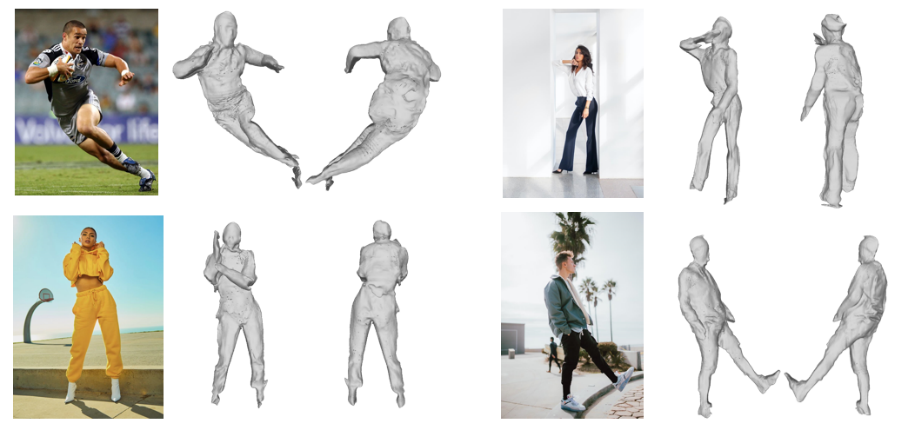}
\end{center}
  \caption{Reconstruction results in some challenging walking, standing, and running poses.}
  \label{challengingpose}
\end{figure}

\begin{figure}[t]
\begin{center}
  \includegraphics[width=0.95\linewidth]{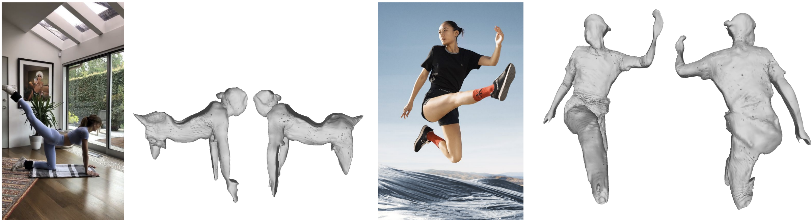}
\end{center}
  \caption{Some failure cases when SMPL fitting is inaccurate.}
  \label{failurecase}
\end{figure}

PIFu~\cite{saito2019pifu} cannot well capture poses due to the depth ambiguity. ICON generates 3D shapes with lumpy surfaces, and the overall shapes are similar to SMPL bodies without clothes.
In contrast, our method generates more realistic avatars with estimated skinning weights and better image fusion.
Besides, our method shows good generalization by reconstructing pants in People-Snapshot even without the clothes in the training set.
The reconstruction results on the challenging poses including walking, standing, and running are illustrated in Fig.~\ref{challengingpose}.
Existing methods for single-image reconstruction often fail in generating a complete human body, resulting in a backside that appears overly smooth. When introducing multi-view images for reconstruction, current methods typically rely on accurate camera calibration, which may not be practical in real-world scenarios. Our proposed method enables the reconstruction of an animatable human body from multiple images with unconstrained poses and cameras, providing a more practical solution.
Besides, We show some failure cases when SMPL fitting is inaccurate in Fig.~\ref{failurecase}.
The primary limitation of our proposed method is the potential risk that inaccurate pose estimation brings errors in mapping 3D points to different image planes for feature fusion. This may cause irrelevant features to be fused, which could potentially deteriorate the quality of the 3D reconstruction.

\begin{figure}
\begin{center}
  \includegraphics[width=0.85\linewidth]{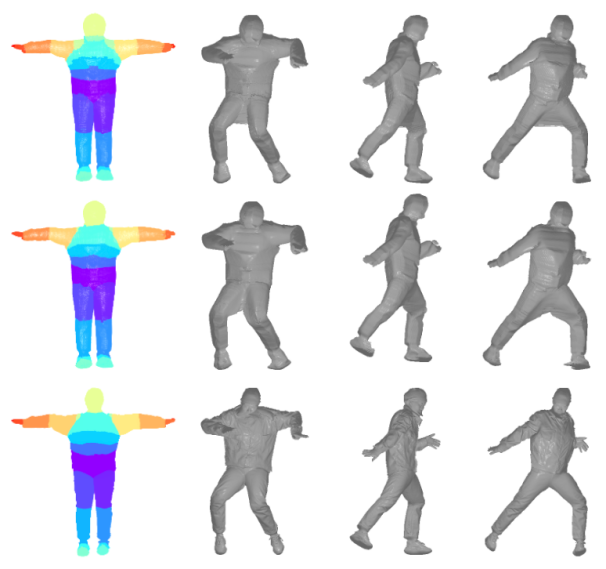}
\end{center}
  \caption{Shapes animated by the nearest neighbor skinning weights (top row), the estimated skinning weights (middle row), and the ground truth (bottom row). Please see the crotch.}
  \label{fig-skin}
\end{figure}

We also evaluate the effectiveness of better skinning weights in animation. As shown in Fig.~\ref{fig-skin}, compared with the common Nearest Neighbor Skinning weights (NN-Skin)~\cite{huang2020arch}, we learn continuous skinning weights which can articulate a mesh smoothly while retaining coherent geometric details.

\subsection{Performance Analysis}
\noindent \textbf{Ablation Study}: We evaluate our method with several alternatives to assess the factors that contribute to the performance. First, with the fitted SMPL, our baseline employs NN-Skin to sample the image features across frames, which are concatenated to regress a pixel-aligned implicit function~\cite{saito2019pifu}. Second, the proposed SKNet is employed to refine the skinning weights. Third, the spatial features (Spatial) extracted from the implicit skinning fields are concatenated to the image features. Finally, the self-attention based feature fusion (FA) module is utilized to improve the robustness of inaccurate fitting and self-occlusion.  As observed in Table~\ref{table-ablation}, each proposed component improves the final performance.

\begin{table}
    \caption{
        Ablation study on MVP-human, evaluated by normal, P2S (cm), and Chamfer error (cm).
        The ``SkinNet'', ``Spatial'', and ``FA'' refer to the skinning weights network, spatial feature, and feature fusion module, respectively. The best results are highlighted. Lower values are better.}
    \label{table-ablation}
    \begin{center}
        \begin{tabular}{ccc|ccc}
            % \hline
             \toprule[2pt]
            \multicolumn{3}{c|}{Components} & \multicolumn{3}{c}{Metrics}  \\
            \hline
            \textit{SKNet} & \textit{Spatial} & \textit{FA}   &  Normal & P2S & Chamfer  \\
            \hline
                      &             &       &   0.0217 &  1.4103 &  1.7981   \\
            \checkmark      &             &      &   0.0232  & 1.3760  & 1.7595  \\
            \checkmark       &   \checkmark        &       &  0.0215 & 1.3455   &  1.7243 \\
            \checkmark       &   \checkmark    &   \checkmark  & \textbf{ 0.0214 } & \textbf{ 1.3418  } & \textbf{ 1.7192 }  \\
            % \hline
            % \hline
            \bottomrule[1.5pt]
        \end{tabular}
    \end{center}
\end{table}

\begin{figure}
  \centering
  \includegraphics[width=0.48\textwidth]{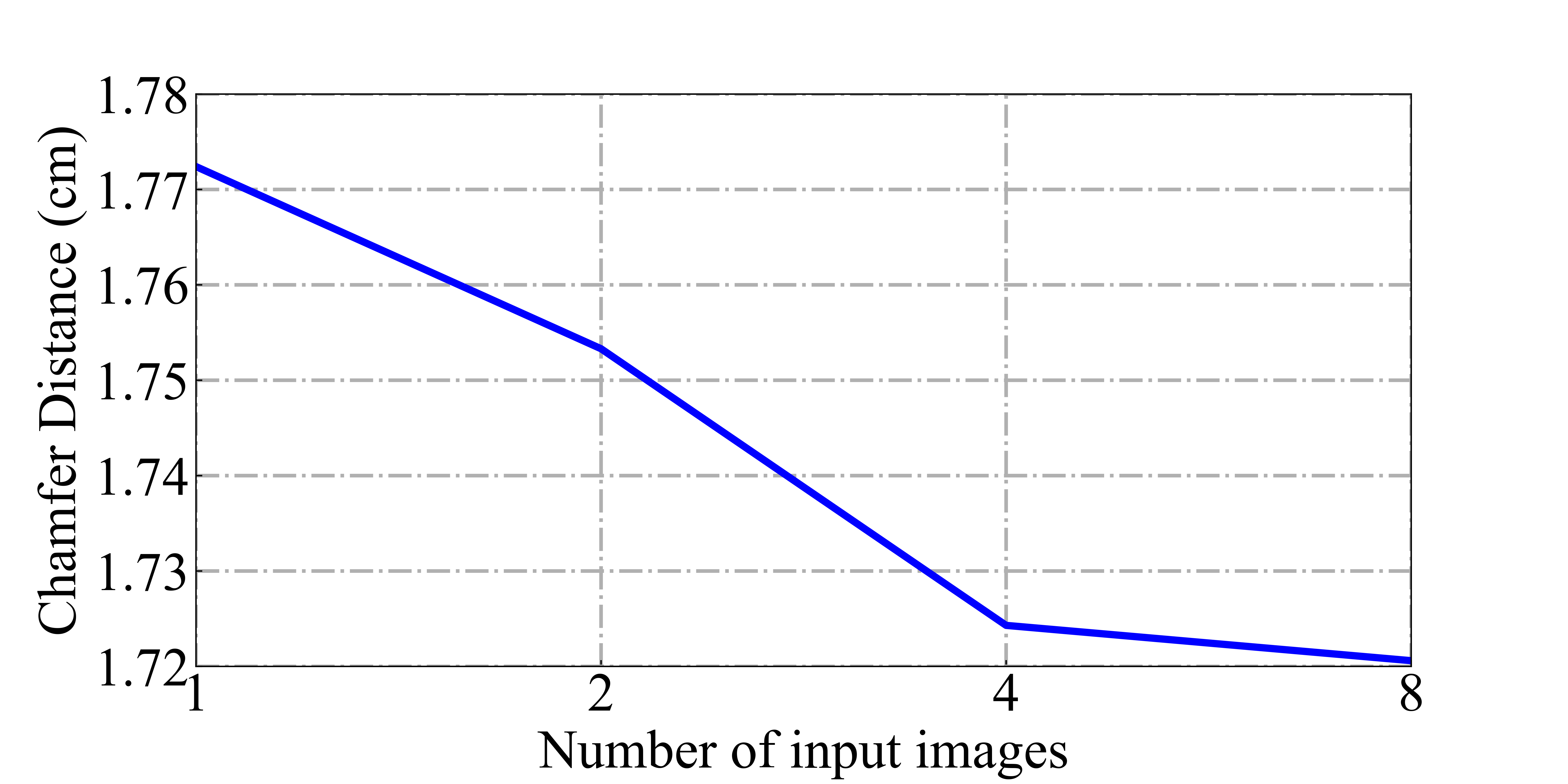}
  \caption{Chamfer errors on MVP-human with different number of inputs. For fast evaluation, we utilize a smaller image encoder.}
  \label{fig-input-num}
\end{figure}

\noindent \textbf{Analysis on Input Number}: Fig.~\ref{fig-input-num} shows the 3D reconstruction errors with the growing number of inputs, from $1$ to $8$. We can see that the performance becomes better as the number of inputs increases and approximately saturates at about $4$ inputs. For efficiency, the $4$-way inputs are utilized in our method.

\begin{figure}
  \centering
  \includegraphics[width=0.48\textwidth]{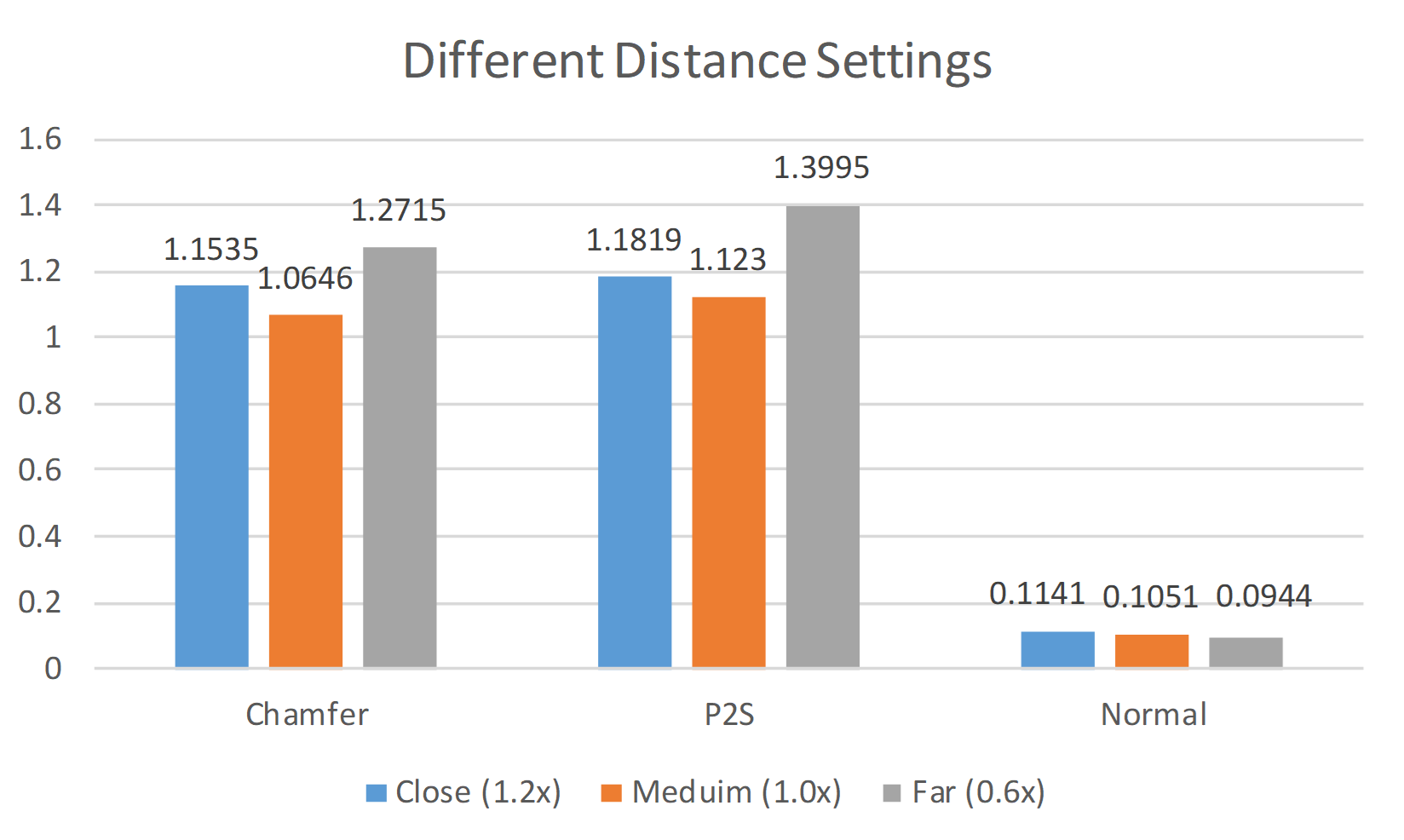}
  \caption{Reconstruction errors in three input distance settings, close (1.2x resolution), medium (1.0x resolution), and far (0.6x resolution).}
  \label{fig-different-distance-settings}
\end{figure}

\noindent \textbf{Analysis on Distance}: We analyze the robustness of inputs from different distances. Considering the main difference brought by distance is resolution, we render the test scans in three distance settings, close (1.2x resolution), medium (1.0x resolution), and far (0.6x resolution), and evaluate the performances.
In Fig.~\ref{fig-different-distance-settings}, we can see that when training and testing share the same distance (1.0x resolution), the performance is the best, followed by close (1.2x resolution) and far (0.6 resolution). The experiments validate that distance has an influence on accuracy.

\section{Conclusion}
In this work, we create a large 3D human dataset MVP-Human with $6,000$  3D scans, and $48,000$ images for $400$ subjects, which is one of the largest 3D human datasets equipped with multi-pose high-quality scans for each subject. We consequently propose a method that introduces the skinning weights network and surface reconstruction network to learn a 3D human clothed avatar from unspecific frames, providing a good baseline performance in this field. Since many previous works are conducted on private datasets, we do hope the publicly available MVP-Human dataset and the baseline method can well promote the development of the avatar reconstruction.

\section*{Acknowledgement}
This work was supported in part by the National Key Research \& Development Program (No. 2020YFC2003901), Chinese National Natural Science Foundation Projects \#62176256, \#62276254, \#62206280, \#62106264, \#62206276, the Youth Innovation Promotion Association CAS (\#Y2021131) and the InnoHK program.

%\noindent \textbf{Limitations and Future Work}: One prerequisite of our method is the pose and camera parameters provided by SMPL fitting.  The parameters cannot be corrected in the current pipeline if the fitting fails. Therefore, the future work will introduce SMPL fitting in the pipeline, and refine pose and camera parameters together with 3D avatar reconstruction.

% \fi
\IEEEpeerreviewmaketitle

\ifCLASSOPTIONcaptionsoff
  \newpage
\fi



%\bibliographystyle{IEEEtran}
%\begin{thebibliography}{1}
%\end{thebibliography}

\bibliographystyle{IEEEtran}
\bibliography{egbib}

\end{document}